\documentclass[sigconf]{acmart}

\AtBeginDocument{%
  }

\setcopyright{acmlicensed}
\copyrightyear{2024}
\acmYear{2024}
\acmDOI{XXXXXXX.XXXXXXX}

\acmConference[MMAsia '24]{ACM Multimedia Asia}{Dec 03-06,
  2024}{Auckland, NZ}
\acmISBN{978-1-4503-XXXX-X/18/06}

\usepackage{subcaption}
\begin{document}

\title{Disentangling Singlish Discourse Particles with Task-Driven Representation}

\author{Linus Tze En Foo}
\affiliation{%
  \institution{University of Edinburgh}
  \city{Edinburgh}
  \country{United Kingdom}}
\email{t.e.l.foo@sms.ed.ac.uk}

\author{Lynnette Hui Xian Ng}
\affiliation{%
  \institution{Carnegie Mellon University}
  \city{Pittsburgh}
  \country{United States}}
\email{lynnetteng@cmu.edu}


\renewcommand{\shortauthors}{Foo and Ng}

\begin{abstract}
Singlish, or formally Colloquial Singapore English, is an English-based creole language originating from the SouthEast Asian country Singapore. The language contains influences from Sinitic languages such as Chinese dialects, Malay, Tamil and so forth. A fundamental task to understanding Singlish is to first understand the pragmatic functions of its discourse particles, upon which Singlish relies heavily to convey meaning. This work offers a preliminary effort to disentangle the Singlish discourse particles \textit{lah, meh, hor} with task-driven representation learning. After disentanglement, we cluster these discourse particles to differentiate their pragmatic functions, and perform Singlish-to-English machine translation. Our work provides a computational method to understanding Singlish discourse particles, and opens avenues towards a deeper comprehension of the language and its usage.
\end{abstract}




\maketitle

\section{Introduction}
Singlish, or more formally known as 
Colloquial Singapore English, is a English-based creole language. Formed from a multicultural society, Singlish contains influences from Sinitic languages such as Hokkien, Cantonese and Mandarin, Malay, Tamil and other varieties in the Singapore language ecology \citep{CSE}. A key feature of Singlish is the presence of a rich and well defined set of discourse particles (i.e. \textit{lah, lor, meh}) which often play multiple pragmatic roles in an utterance \citep{gupta_1992,multifunction}. Therefore, a fundamental Natural Language Processing (NLP) task is differentiating the pragmatic functions of the discourse particles; that is, identifying the meaning that the speaker intended to convey to the listener with the particle \citep{discourseparticles}.

In doing so, the key difficulty of Singlish language processing is the multi-functionality of its particles. While previous works \citep{singlish_platt,Wee_2018} have attempted to define the function of some particles, the definitions differ due to different particle functions in different contexts \citep{gupta_1992}. To add to the difficulty, a particle may play a combination of roles in an utterance. For example, \textit{meh} in ``He fainted \textit{meh}?" simultaneously expresses disbelief and surprise \citep{discourseparticles}.

Past work in the Singlish language largely involved linguistic discussions on its structure and particle usages \citep{singlish_platt,gupta_1992,discourseparticles}. Within the computational realm, \citet{sms_corpus} curated short messages written in multiple languages, of which one is Singlish, and this dataset has been used for tasks such as text clustering and spam detection \citep{parimala2012study,almeida2011contributions}. \citep{chen2015corpus} created curated the pronunciation variations for Singlish, providing a data-driven transcription of the differences in pronunciation between Singlish and English.

\begin{sloppypar}
This work aims to disentangle Singlish discourse particles through task-driven representation learning, revealing the pragmatic functions of the particles. Particle representation thus lead us to two tasks: clustering of pragmatic functions of the particles, and Singlish To English machine translation. This work presents the potential to be exploited in downstream tasks that rely heavily on sentence semantics like stance identification. Particularly, we perform the following NLP tasks:
\end{sloppypar}
\begin{itemize}
    \item \textbf{Task-Driven Representation of Singlish Discourse Particles}. We encode the nuanced differences in the usages between and within the particles using two training tasks: Next Sentence Prediction and Particle Prediction.
    \item \begin{sloppypar}
    \textbf{Clustering of Pragmatic Functions of Singlish Discourse Particles.} We perform unsupervised clustering of the particles' pragmatic functions, showcasing how particles differentiate and overlap in their functional use. 
    \end{sloppypar}
    \item \textbf{Singlish to English Machine Translation.} Even though Singlish is closely related to English, many native English speakers are unable to understand Singlish. We construct a GPT model, GPT-Singlish, on Reddit and forum texts.
\end{itemize}

\section{Task-Driven Representation of Singlish Discourse Particles}
Representation learning algorithms encode latent features of data as contextual embeddings, aiding in downstream tasks. Encoding the Singlish language involves two considerations: (1) the embeddings must tackle polysemous words, or words that have different meaning based on their context \citep{polysemy}; and (2) Singlish is a low-resource language that current models are not geared to handle. 

BERT is a transformer-based language model that is designed to pretrain deep bidirectional representations \citep{vaswani2017attention}. It is a common model used to handle representations of polysemous words and low-resource languages. A pre-trained BERT model performs well on syntactic tasks, signalling its effectiveness in capturing syntactic information in its sentence embeddings. The BERT model has been used on many low-resource language understanding across languages from Bangla to Maltese \citep{bhattacharjee2021banglabert,micallef2022pre} with reasonable success.
Multilingual BERT has been extended to perform Named-Entity Recognition in 27 languages, some of which are low-resource languages \citep{wang2020extending}. 

\subsection{Training a Singlish Representation}
We used a task-based learning approach to train a Singlish representation for three discourse particles: \textit{lah, meh, hor}. That is, we defined two tasks as the training objective of a model. These two tasks are: Next Sentence Prediction (NSP) and Particle Prediction (P-Pred).

\subsubsection{NSP Training Objective} 
\label{nsp}
Training a model with NSP task can improve implicit discourse relation classification \citep{nsp_implicitdiscourse}. We guide the learning of semantics of particles using the Utterance Relevance Classification (URC) metric, as detailed in the BERT-FP implementation \citep{han-etal-2021-fine}.

Each training example is expressed as $(D_i, S^j_i, L_i)$. $D_i$ represents three consecutive multi-turn dialogue such that the final utterance is enforced to have a sentence-final particle. $S^j_i$ represents a candidate next-sentence and $L$ is the label, indicating if $S^j_i$ is ``next", ``random" or ``related". During training, we set $j = 2$ with the ratio of positive (next) to negative (random or related) candidate next-sentence to be 1:1. For each negative example, a related or random candidate sentence is sampled with equal chance along with its corresponding label, $L$. During testing however, we set $j=10$ and the ratio of positive to negative candidate next-sentence to be 1:9. 

\subsubsection{P-Pred Training Objective} 
The next-word prediction task requires the model to encode semantic information of the target sentence\citep{merrill2024can}. Our P-Pred training objective is a variation of this task to predict one of the three sentence-final particles. Each of the three particles are down-sampled to ensure a balanced dataset. Each training example can be formulated as $(d_{i-2}, d_{i-1}, d_i^{'})$ where $d_j$'s refers to sequential dialogues and $d_j^{'}$ refers to the utterance with the sentence-final particle masked out.

\subsection{Data}
We merged two corpora for this study: (1) a subset of the National Speech Corpus (NSC), which consists 2000 hours of orthographically transcribed read speeches, reflecting the Singlish accent and words \citep{nsc}; and (2) the subset of NUS-SMS containing short text messages in Snglish, reflecting the everyday use of the language. This combination of datasets provided diversity in the usage of discourse particles in both speech and writing.

\subsection{Model Construction}
\begin{figure}
    \centering
    \includegraphics[width=\linewidth]{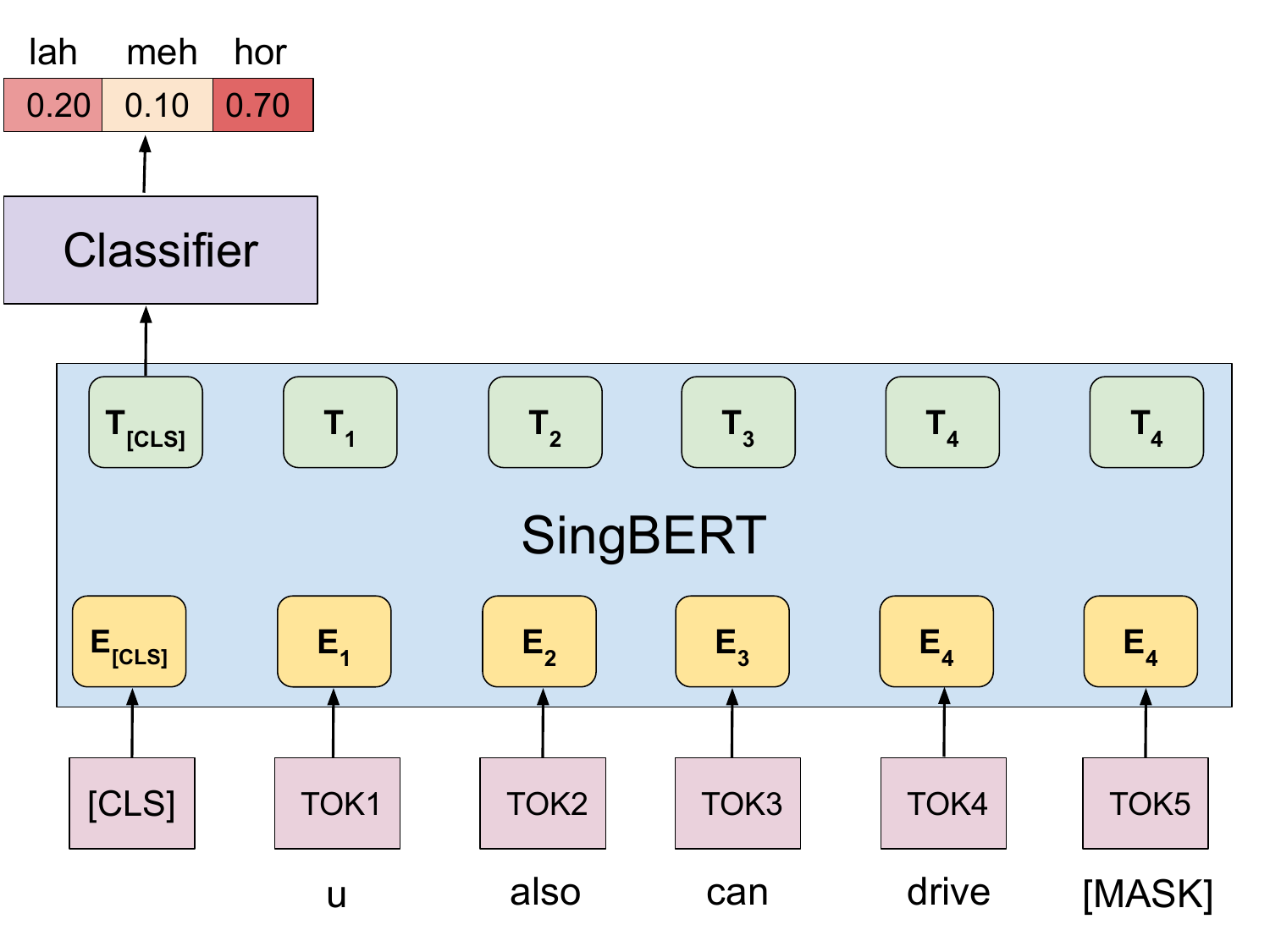}
    \caption{Illustration of model architecture for P-Pred task. The output representation of the \texttt{[CLS]} token from SingBERT is passed into a fully connected classifier to predict the probability for one of the three discourse classes.}
    \Description[A classifier is stacked on top of a SingBERT model]{Weights from the CLS token is passed into the classifier to produce probability for each of the three classes.}
    \label{fig:model-architecture}
\end{figure}

Here is how we constructed a model to train on the NSP and P-Pred tasks. \autoref{fig:model-architecture} visually illustrates the model architecture for the P-Pred task.

\paragraph{Baseline Model} We adopted SingBERT \citep{singbert}, a variant of BERT, which was fine-tuned with the task of Masked-Language-Modelling on a text corpus collected from Singapore-related subreddits and Singapore-based forums. 

\paragraph{Stacking Classifiers} We used SingBERT as the base encoder and initialised it with the same weights as the original authors suggested.  The embedding obtained at the \texttt{[CLS]} token from the output of SingBERT, which is then passed into a classifier. This classifier contains three-fully connected ReLU layers with 768 hidden units each and dropout of 0.2. The classifier outputs the predicted probability of each class. 

\subsection{Evaluation} 
\paragraph{NSP evaluation} For evaluation of the NSP model, we used the recall metric, a standard evaluation measure in retrieval-based systems with imbalanced data \citep{raghavan1989retrieval}. Specifically, $R_N@k$ is used to measure if the correct answer exists among the top k candidates out of N candidate responses. The $R_{10}@1$ on NSP model is 0.483. 

\paragraph{P-Pred evaluation} The P-Pred task presents a balanced dataset, hence we used the accuracy metric. The accuracy for the P-Pred model is 0.790.

\paragraph{NSP on P-Pred}
We further evaluated how the NSP model performs on P-Pred. We thus created a modified test set with a similar formulation. In this test data, we set $j=3$, that is, for each example where the positive candidate next-sentence has a sentence-final particle, we substitute the particle with the other two particles.
The $R_3@1$ on NSP model evaluating on P-Pred results is 0.767. 

\section{Clustering of Pragmatic Functions of Discourse Particles}
Beyond representing discourse particles, we next need to understand what the particles are learning. Hence, we attempt to isolate the representations of each of the three particles, then perform unsupervised clustering on the representations.

\subsection{Data}
We used the representations obtained from the models in the previous section as input data towards the clustering algorithm. Both representations were considered in this section.

\subsection{Linear-Isolation of Representation}
To isolate the representations of the particles, we introduce a technique called Linear-Isolation of Representation (LIR). This technique leverages on the linear-characteristics that particle embeddings exhibit.
Formally, the representation of the particle, $\vec{R}$, in the $i^{th}$ utterance is:
\begin{align}
   \vec{R_i} =  \vec{V_i} - \vec{V_i}^{'}
\end{align}
where $\vec{V_i}$ denotes the original embedding (word or sentence) and $\vec{V_i}^{'}$ denotes the embedding where the particle token is replaced with a \texttt{[MASK]} token. 

\begin{figure*}
\centering
        \includegraphics[width=0.85\linewidth]{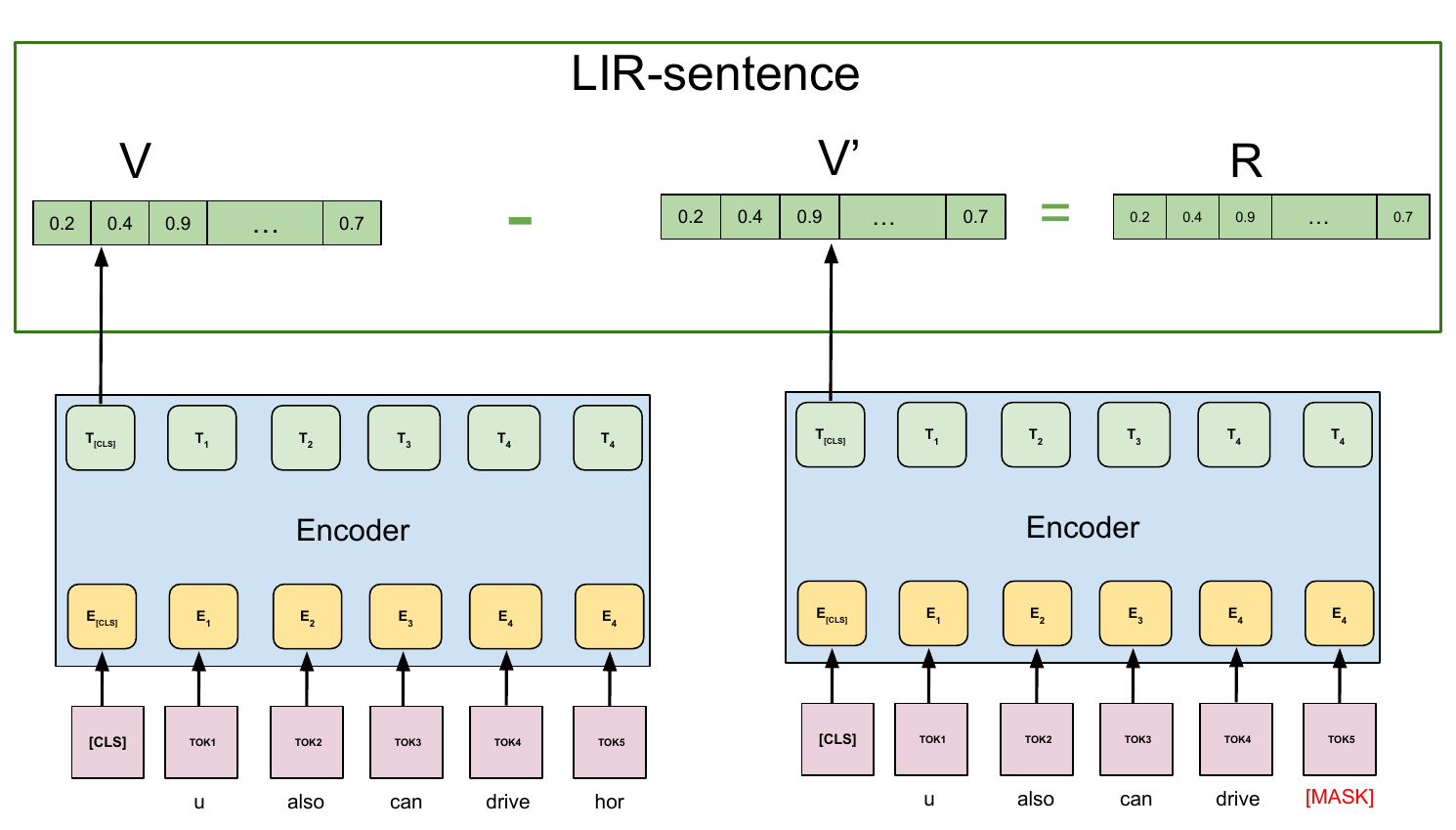}
\caption{Architecture for obtaining word representation using LIR-sentence. LIR-word uses a similar architecture except it uses particle word embedding instead of sentence embedding.}
\label{fig:LIR}
\Description[The method of obtaining the representation using LIR-sentence]{Using LIR-sentence, the representation can be obtained by subtracting sentence embeddings obtained from the encoders.}
\end{figure*}

\begin{figure*}
\centering
\begin{subfigure}{0.3\textwidth}
  \includegraphics[width=\textwidth]{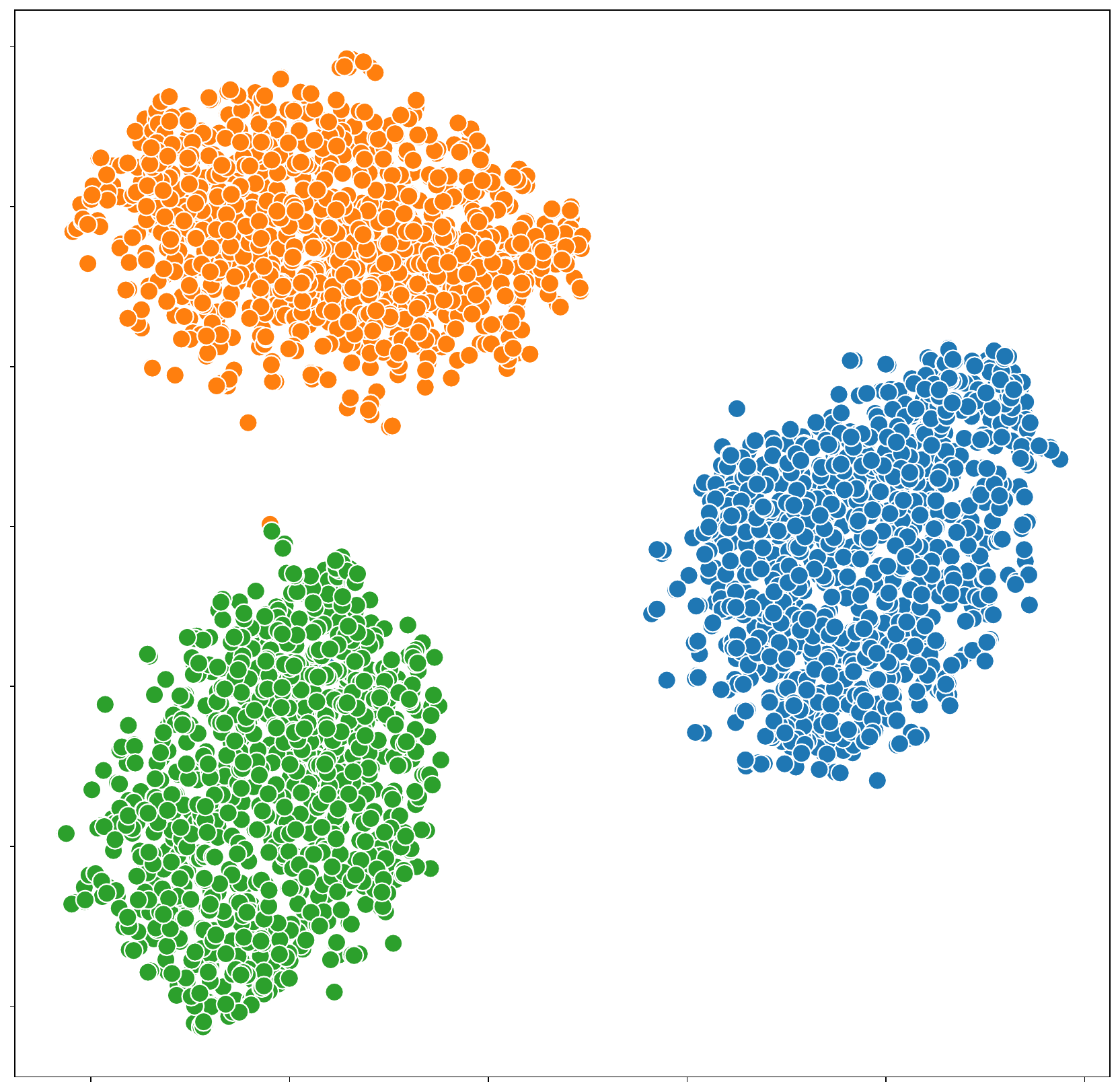}
  \caption{Using LIR-word}
  \label{fig:sub1_1}
\end{subfigure}%
\begin{subfigure}{0.3\linewidth}
  \includegraphics[width=\textwidth]{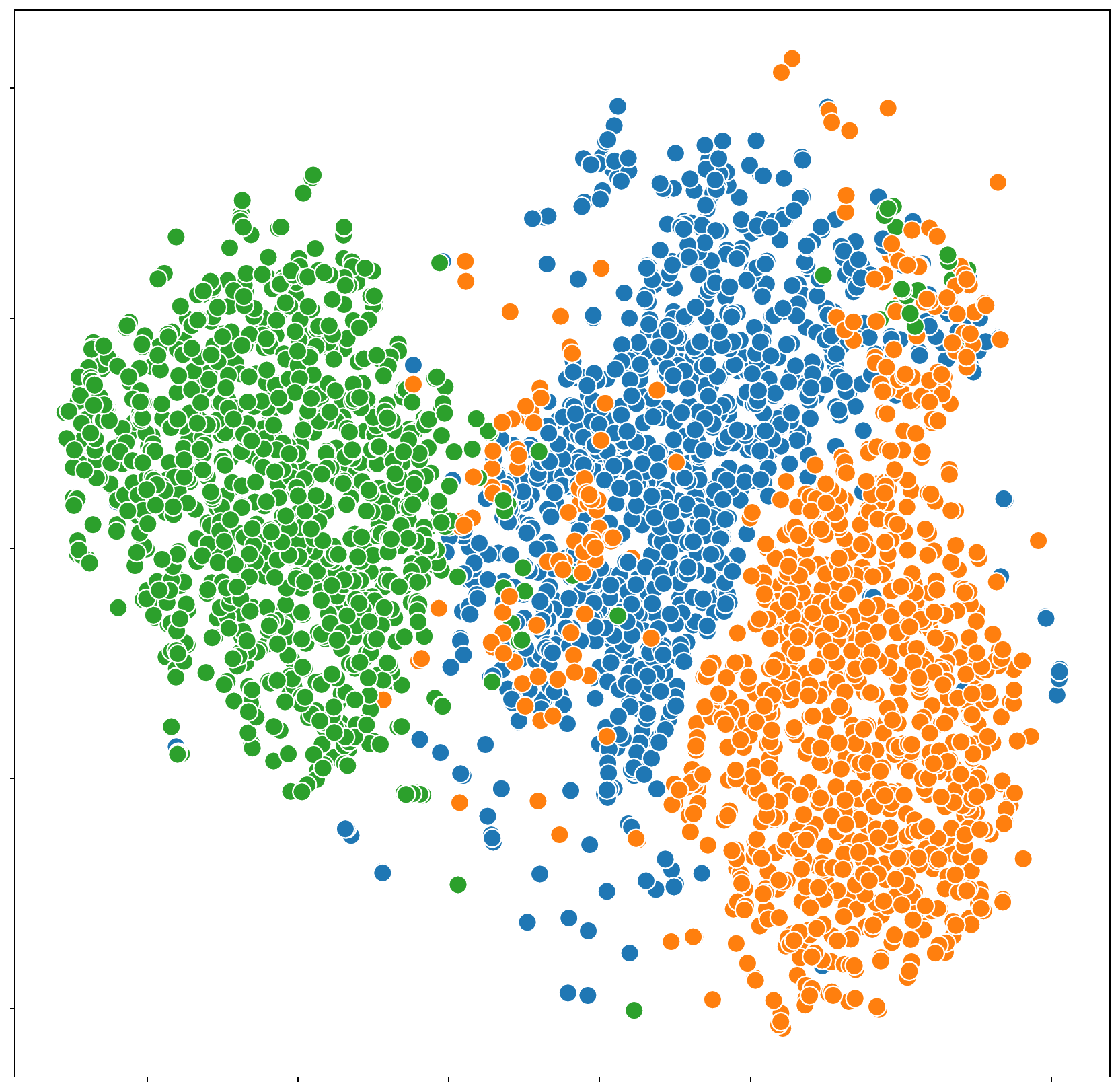}
  \caption{Using LIR-sentence}
  \label{fig:sub2}
\end{subfigure}
\begin{subfigure}{0.3\textwidth}
  \includegraphics[width=\textwidth]{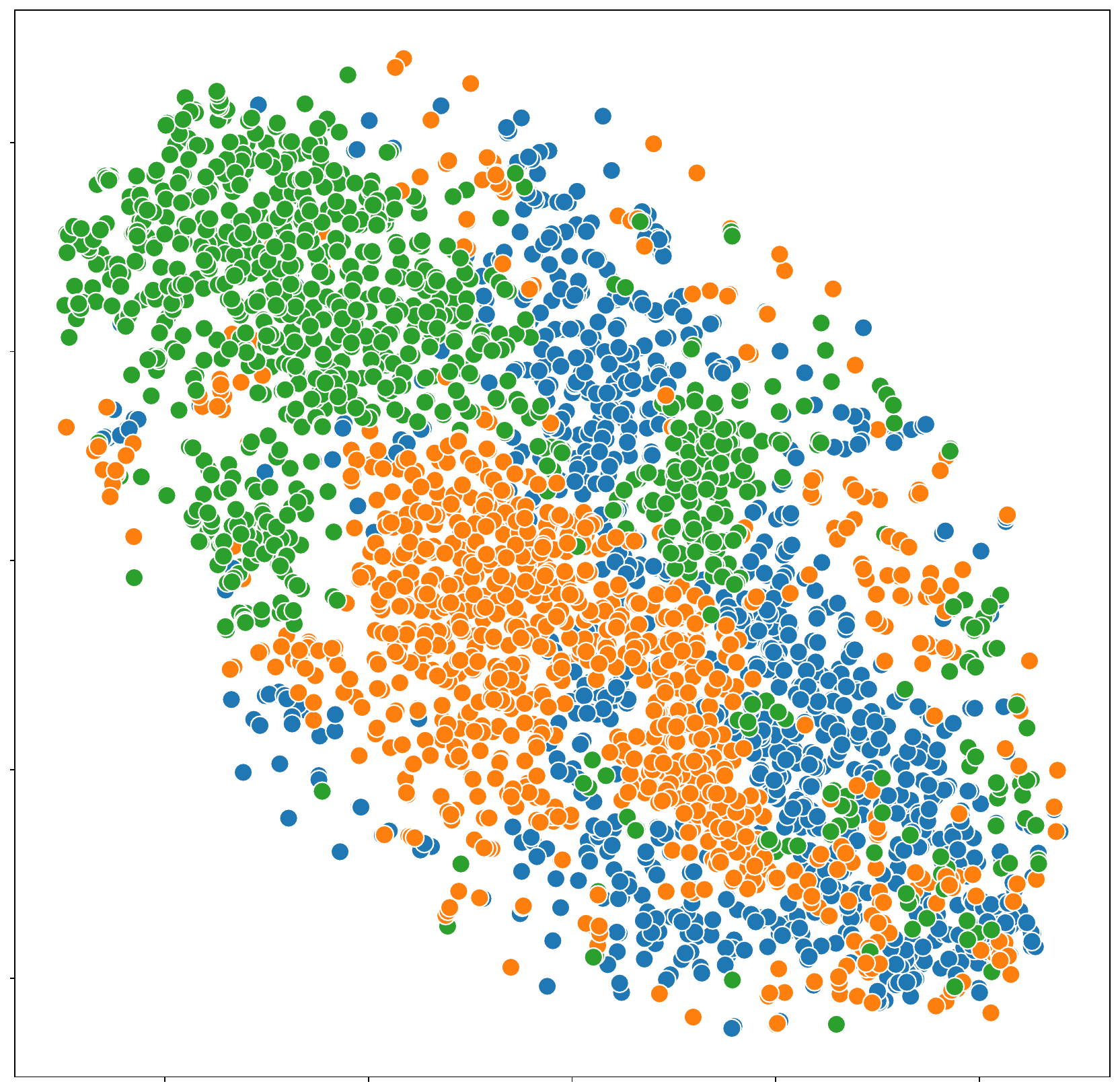}
  \caption{Naive Word-Embedding}
  \label{fig:sub2_1}
\end{subfigure}
\caption{Representation for baseline model SingBERT. Blue - \textit{lah}, orange - \textit{meh}, green - \textit{hor}.}
\label{fig:lir_word_cluster}
\Description[scatter plots of the representations from the different isolation methods]{Using LIR-word the inter-particle clusters are visible whereas for the other methods the particle clusters are overlapping}
\end{figure*}

We experimented with two methods of obtaining representation while adopting the principle of LIR: LIR-word and LIR-sentence. In LIR-sentence, $\vec{V_i}$ and $\vec{V_i}^{'}$ were obtained from the final layer of the respective Bert-based encoders at the position corresponding to the \texttt{[CLS]} token. This token is assumed to be the sentence-embedding. LIR-word is similar except that $\vec{V_i}$ and $\vec{V_i}^{'}$ were obtained at the sentence-final position corresponding to the particle. \autoref{fig:LIR} illustrates the LIR method. These two methods result in slightly different representations, which we illustrate using the baseline model in \autoref{fig:lir_word_cluster}.

\subsection{Unsupervised Clustering}
\begin{figure}[H]
\centering
\begin{subfigure}{0.5\linewidth}
  \includegraphics[width=\textwidth]{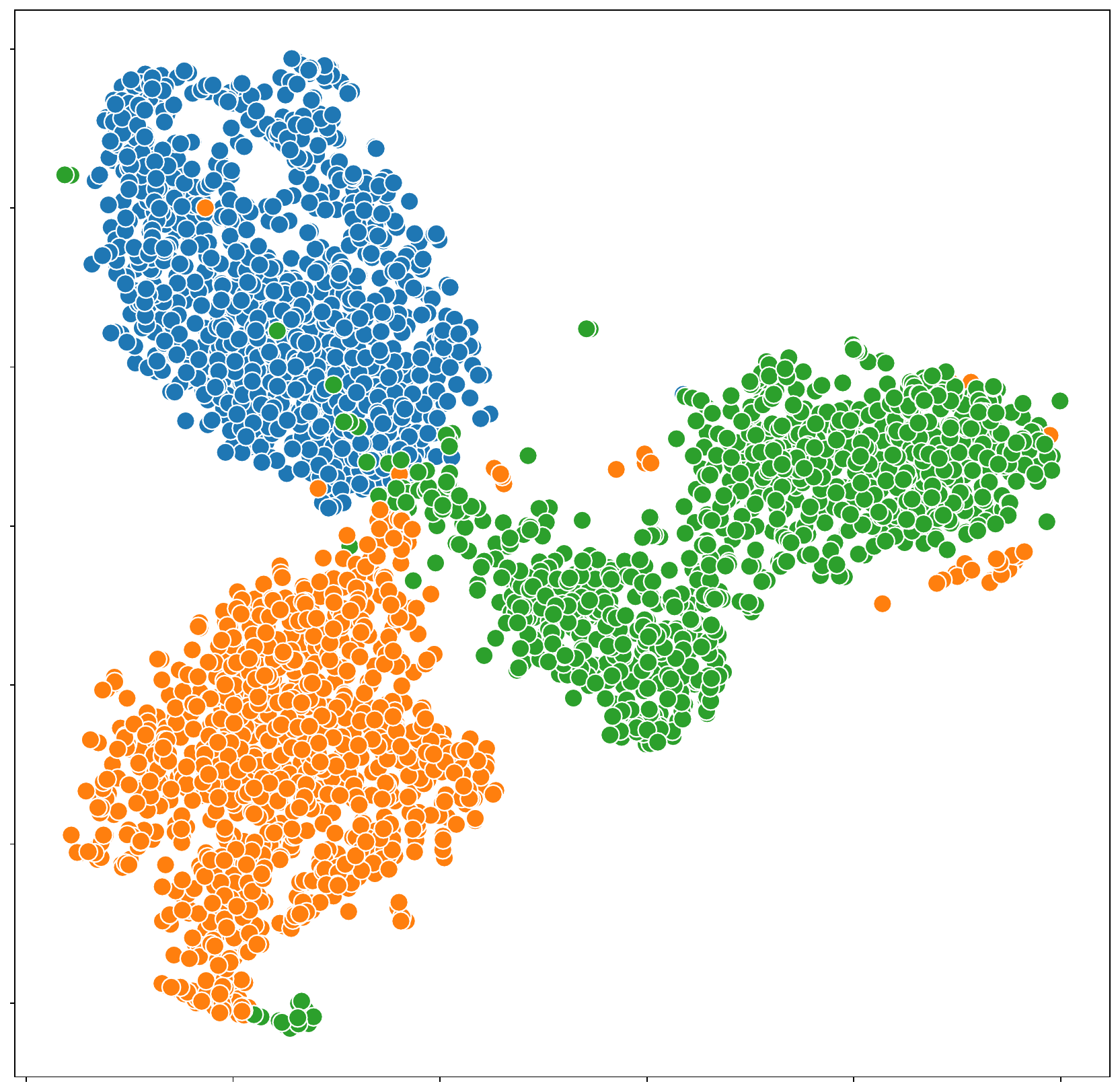}
  \caption{NSP}
  \label{fig:sub1}
\end{subfigure}%
\begin{subfigure}{0.5\linewidth}
  \includegraphics[width=\textwidth]{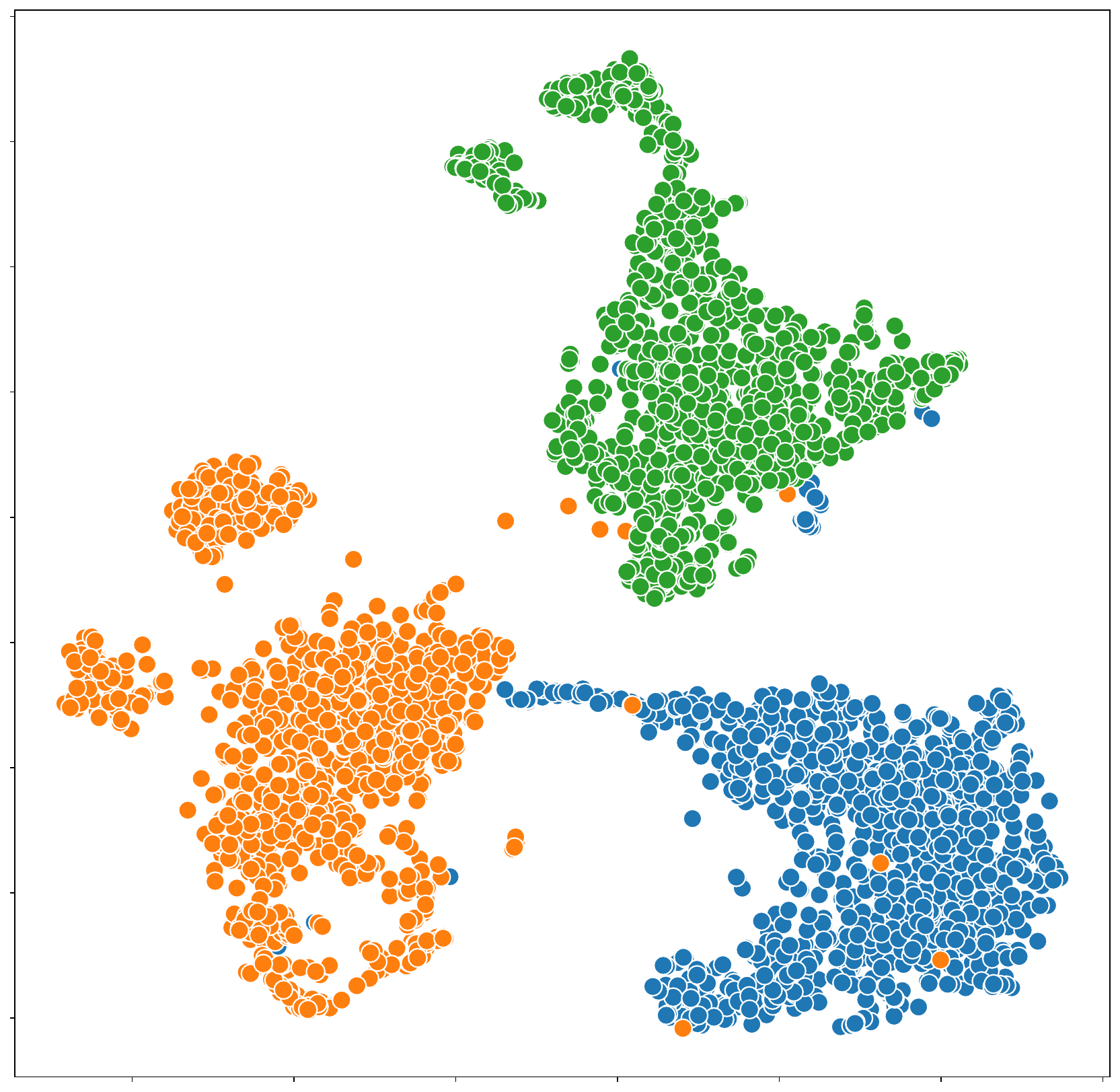}
  \caption{P-Pred}
  \label{fig:sub2_2}
\end{subfigure}
\caption{Clusters of particle representations. Blue - \textit{lah}, orange - \textit{meh}, green - \textit{hor}.}
\label{fig:rep-visualisation}
\Description[particles are visibly scattered with clear inter-particle clusters]{particles are visibly scattered with clear inter-particle clusters for both NSP and P-Pred model. For the P-Pred model, additional intra-particle clusters can be seen}
\end{figure}

To understand the natural aggregation of representations, we perform unsupervised clustering on the representation vectors. The clusters reflect the pragmatic functions of the discourse functions. We use the t-SNE method \citep{tsne} for dimensionality reduction of the representations, before plotting them in a two-dimensional plot. We then perform unsupervised clustering with the DBSCAN method. The parameters of the DBSCAN were adjusted to achieve the highest silhouette score, a measure of the overall quality of clustering, while ensuring that there is at least one cluster corresponding to each cluster. The plots are visualised in \autoref{fig:lir_word_cluster} and \autoref{fig:rep-visualisation}. Results of the silhouette scores from the clusters are presented in \autoref{tab:table_results_sil}.

\begin{table}
\centering
\resizebox{0.47\textwidth}{!} {
\begin{tabular}{|c|c|c|c|}
\hline
\textbf{Model} & \textbf{Naive-Word Embedding} & \textbf{LIR-word} & \textbf{LIR-sentence} \\ \hline
Baseline & -0.0224  & \textbf{0.436}  & 0.379 \\
P-Pred  & 0.032  & 0.240    &  \textbf{0.395}         \\
NSP & 0.275  & \textbf{0.413} &  0.311        \\
\hline
\end{tabular}
}
\caption{Silhouette scores of clusters. \textbf{Bold} scores represent best performing representation for each model.}
\label{tab:table_results_sil}
\end{table}

\begin{sloppypar}
Based on the silhouette scores obtained for each representation method, we then visualise the representation obtained from the best performing method which in this case is LIR-word for the Baseline and NSP model, and LIR-sentence for P-Pred. The visualisation can be seen from Figure \ref{fig:rep-visualisation}.
\end{sloppypar}

\subsection{Interpreting Clusters}
On top of visually inspecting the presence of clusters, we quantitatively interpreted some clusters. We manually curated a set of particle pragmatic function (summarised in \autoref{tab:function_table}). This set selects for two functions for each particle, chosen for functions that are as mutually exclusive as possible, to maximise the separation between clusters.

We then manually labeled the pragmatic function of these particles in a random sample of 40 clustered texts. We then used the Adjusted Rand Index (ARI) the similarity between the labeled data and the unsupervised clustering labels, presented in \autoref{tab:ARI_table}. The ARI is a statistical measure ranging between 0 and 1 that evaluates the similarity between two clusters. A value closer to 1 indicates a perfect match between the clusters, while a value close to 0 indicates a random assignment of points to clusters. For both the NSP and P-Pred task, there is significant evidence of function-based clustering for \textit{meh} and \textit{hor} particles ($p<0.01$), while there is a lack of inter-particle clustering for the \textit{lah} particle.

\begin{table}
\centering
\begin{tabular}{|c|c|}
\hline
\textbf{Particle}  & \textbf{Functions} \\ \hline
\textit{lah} &  Friendliness,
Hostility  \\\hline     
\textit{meh} &    Disagree, Expresses surprise  \\ \hline     
\textit{hor} &   Reduce Harshness, Solicit agreement \\ \hline 
\end{tabular}
\caption{Selected subset of pragmatic functions}
\label{tab:function_table}
\end{table}

\begin{table}
    \centering
    \begin{tabular}{|c|c|c|c|c|c|c|}
    \hline
        \textbf{Task} & \multicolumn{3}{|c|}{\textbf{\# of clusters}} & \multicolumn{3}{|c|}{\textbf{ARI values}} \\ \hline 
        ~ & lah & meh & hor & lah & meh & hor \\ \hline 
        Baseline & 1 & 1 & 1 & 0.0 & 0.0 & 0.0 \\ \hline 
        NSP & 1 & 4 & 3 & 0.0 & 0.091*** & 0.142*** \\ \hline 
        P-Pred & 1 & 3 & 4 & 0.0 & 0.151*** & 0.194*** \\ \hline 
    \end{tabular}
    \caption{Analysis of clusters formed in representing particles. *** indicates significance at $p<0.01$ level}
    \label{tab:ARI_table}
\end{table}

\begin{table}
    \centering
    \begin{tabular}{|c|p{6cm}|}
        \hline 
        \textbf{Cluster} & \textbf{Example utterance} \\ \hline 
        Cluster 1 & *really \textit{meh} \\ \hline
        Cluster 2 & 1. but we drinking at a bar no \textit{meh} \newline 
                    2. good what no \textit{meh} \newline 
                    3. still not allowed in singapore what no \textit{meh}
                    \\ \hline 
        Cluster 3 & *no \textit{meh} \\ \hline 
    \end{tabular}
    \caption{Examples of utterances in three clusters for \textit{meh}}
    \label{tab:cluster_example}
\end{table}

\paragraph{Discussion}
By both qualitative visual inspection and quantitative sihouette scores, our LIR technique is superior to traditional word-embedding methods. However, the effect of using LIR-word and LIR-sentence varies according to the representation training. We posit that P-Pred performs better in LIR-sentence than in LIR-word because the particle is masked out at the training step to facilitate prediction, reducing the model's ability to learn contextualised word embeddings to encode its pragmatics. In the NSP model, however, all particles are present during training, thus the model performs better at the word-embedding level.

Regardless of model, the clustering techniques fail to form clusters for the particle \textit{lah}. We hypothesise that the functions performed by \textit{lah} in most contexts are not mutually exclusive, resulting in the inability to disentangle its individual function. We leave this for future studies to better understand this observation.

Even though the ARI values provide evidence of pragmatic-based clustering, there is evidence of similar words present within each cluster (see \autoref{tab:cluster_example}). There is possibly word-based clustering on top of pragmatic-based clustering, illustrating that phrases and clauses containing a particle can be associated with pragmatic functions. For example, the phrase \textit{really meh} signals surprise while \textit{no meh} indicates disagreement. Future work could benefit from correlating the phrase and feature particles.

Lastly, we compare the inter-relation of the silhouette scores and the ARI. While the silhouette scores obtained from the best performing representations of both P-Pred and NSP are similar to that of the baseline model (fine-tuned SingBERT), the ARI reveals that the baseline fails to firm any inter-particle function clustering, and is only capable of intra-particle clustering. Therefore, our NSP and P-Pred representation models are successful in forming inter-particle clustering of discourse particles based on pragmatics. 

\section{Singlish-to-English Machine Translation}
Translation of Singlish-to-English creates accessibility to the language. Unlike typical language translation, Singlish as a creole contains more noise, e.g. varying spellings of the same word. Singlish-English data is a low-resource translation, which presents an opportunity to explore the effectiveness of current Machine Translation (MT) techniques.
\citet{liu2022singlish} performed Singlish-to-English translation using a linguistic hierarchy method, by working on three subtasks: lexical level normalization, syntactic level editing, and semantic level rewriting. \citet{Chakraborty_2024} performed English-to-Singlish translation by fine-tuning the H2O Danube2 model, and demonstrated capabilities in generating fluent Singlish texts.

\subsection{Data}
We collected a total of 80,000 Singlish texts from Singapore-related subreddits and the Singapore-based forum HardwareZone. A subset of 400 Singlish texts are translated to English by the authors, who are native Singlish speakers. This translated subset is termed \textit{manual-subset-eng}. We thus use these texts as training data to create a machine translation named GPT-Singlish.

\subsection{Model Construction}
\paragraph{Initial Model Construction}
We adapt the GPT-2 model because of its small size, and it is less data-hungry model. We perform translation with the following prompt: $<$Singlish$>$ \{singlish\_text\} $<$English$>$." Our preliminary visual investigation shows that while there are examples that are correctly translated (\autoref{tab:translation_right}), majority of the messages were incorrectly translated by the machine (\autoref{tab:translation_wrong}). Such results indicate that GPT-2 had not been trained on much Singlish-English related data, making it an ideal baseline model to improve upon.

\paragraph{Back-Translation}
Singlish is a low-resource language, which lacks quality parallel data to train MT models. As such, we use back-translation, an iterative data augmentation technique to increase the size of the training data \citep{backtranslation}.

To improve the machine-translation, we trained a smaller statistical translation model, Moses \citep{moses}. We used the 400 annotated sentences to act as a ``forward model" in this portion, which returned 4,500 synthetic parallel Singlish-to-English data. We trained 2 iterations of back-translation, achieving a BLEU score of 65.

\subsection{Model Evaluation}
\begin{sloppypar}
At initial glance, a BLEU score of 65 is promising. However, upon closer visual inspection, the authors noted that the machine-translation were inaccurate, and that inflated BLEU score could be due to the significant overlap in syntax and lexicon between both source and target languages.
\end{sloppypar}

We thus performed sentiment analysis to investigate the model's ability to preserve semantics through translation. We randomly selected 100 English texts from \textit{manual-subset-eng}. We obtained the sentiment of the English texts using the FastText library \citep{bojanowski2016enriching} (\textit{sent-eng}). We pass the Singlish counterpart of the texts into GPT-Singlish to obtain \textit{auto-subset-eng}, which are machine-translated English texts. We next evaluate the sentiment of \textit{auto-subset-eng} with FastText. Ideally, both sentiments of \textit{manual-subset-eng} and \textit{auto-subset-eng} should match. However, we obtain an accuracy of 68.1\%. The accuracy decreases to 66.7\% if translation is not performed. This investigation shows that our translation step helps to sentence semantic, but can still be improved.

\paragraph{Discussion} GPT-Singlish model performed with a high BLEU score, indicating the possibility of using machine translation to decode the language for a wider audience. However, perception-based evaluation by native Singlish speakers noted that many sentences were incorrectly translated. Our extrinsic evaluation shows that our model improves on semantically similar translation. This observation suggests that current evaluation methods may be inadequate for creoles because syntax and lexicon overlap with English, yet are sematically different. Further work can expand on the accuracy of Singlish machine translation and develop better evaluation frameworks for creole languages. Ideas include adopting transfer learning from models trained on related languages \citep{close-realated-mt}.

\begin{table*}
\begin{tabular}
{|p{5.5cm}|p{5.5cm}|p{5.5cm}|}
\hline
\textbf{Singlish} & \textbf{Machine Translation} & \textbf{Manual Translation}  \\ \hline
cham liao... we copy theirs sia... how ar &
that this... we them no.. what ar & 
oh no... we copy theirs... what do we do \\ \hline
better than talking like some low class ah beng ma &
better than talking kena moi this low class ah beng ah &
better than talking like some low class gangster \\ \hline
teacher call your mother, alamak how &
teacher call your mother, alamak how &
teacher call your mother, that is not good, how did it go \\ \hline
\end{tabular}
\caption{Example of poorly machine-translated texts}
\label{tab:translation_wrong}
\end{table*}

\begin{table*}
\begin{tabular}
{|p{5.5cm}|p{5.5cm}|p{5.5cm}|}
\hline
\textbf{Singlish} & \textbf{Machine Translation}  & \textbf{Manual Translation} \\ \hline
he is a real eh sai investor & he is a real good investor & he is a really good investor \\ \hline
I buay tahan the boss’s jokes already &
I cannot stand the boss's jokes already &
I cannot stand the boss's jokes already \\ \hline
now still got people play card game meh &
now do people still play card game &
do people still play card game nowadays \\ \hline
\end{tabular}
\caption{Examples of correctly machine-translated texts}
\label{tab:translation_right}
\end{table*}

\section{Conclusion}
Singlish is an important language to study, because it is the colloquial language of a country with the highest GDP in southeast asia, and of 5.6 million people as of 2022.
This work contributes to the understanding of the language with a systematic methodology to understand the core of the language -- Singlish discourse particles. We disentangle these particles with task-driven representation methods, cluster the particles and perform a machine translation task. 
Our explorations, while varied, focuses mostly on the orthographic form of particles and does not consider tonal variation. However, many of the intra-particle functionalities can be easily differentiated with additional tonal information. Deeper investigations can be performed to integrate tonal information, which could improve the pragmatic representations of these particles.
Nonetheless, we hope our explorations and methodology  facilitates understanding of the language.

\begin{acks}
We thank Dr. Peter Bell for his guidance and the University of Edinburgh for providing access to computational resources.
\end{acks}

\bibliographystyle{ACM-Reference-Format}
\bibliography{sample-base}

\appendix

\end{document}